\documentclass{llncs}

\usepackage{graphicx}
\usepackage{xcolor}
\usepackage{cite}
\graphicspath{{./images/}}
\DeclareGraphicsExtensions{.pdf,.jpeg,.png}

\usepackage{amsmath}
\usepackage{amssymb}

\usepackage{array} 

\usepackage[noend]{algpseudocode}
\algrenewcommand\algorithmicrequire{\textbf{Precondition:}}
\algrenewcommand\algorithmicensure{\textbf{Postcondition:}}
\algdef{C}[IF]{IF}{ElsIf}[1]{\textbf{elif} #1\ \algorithmicthen}
\algnewcommand{\LeftComment}[1]{\Statex \(\triangleright\) #1} 
\usepackage{acl2}

\algrenewcommand\algorithmicindent{1.0em}

\newcommand{\codestyle}[1]{\fontfamily{lmtt}\bfseries\selectfont #1}
\DeclareTextFontCommand{\code}{\codestyle}

\newcommand{\namef}[1]{\ensuremath{\textsc{#1}}}
\newcommand{\varf}[1]{\ensuremath{\mathit{#1}}}
\newcommand{\nil}{\code{nil}}
\newcommand{\True}{\code{true}}
\newcommand{\False}{\code{false}}
\newcommand{\Proved}{\code{proved}}
\newcommand{\Unknown}{\code{unknown}}
\newcommand{\assign}{:= \,}
\newcommand{\ie}{\emph{i.e.}}
\newcommand{\eg}{\emph{e.g.}}
\newcommand{\etal}{\emph{et al.}}

\algblockdefx[structure]{Structure}{EndStructure}[1][]{\textbf{structure} #1}{}

\makeatletter
\ifthenelse{\equal{\ALG@noend}{t}}%
  {\algtext*{EndStructure}}
  {}%
\makeatother
\algdef{SL}[ALIAS]{Alias}{0}[2]{\textbf{alias} \varf{#1} \assign \varf{#2}}

\newcommand{\fieldf}[1]{\textsf{#1}}
\newcommand{\field}[2]{\State \fieldf{#1} \Comment #2}

\newcommand\TEMPALGOCAPTION{}
\newcommand\TEMPALGOLABEL{}
\newenvironment{myalgorithmic}[2]{
  \begin{figure}[bt]
    \def\TEMPALGOCAPTION{#1}
    \def\TEMPALGOLABEL{#2}
    \small
    \begin{algorithmic}[1]
  }{
    \end{algorithmic}
    \normalsize
    \caption{\TEMPALGOCAPTION}
    \label{\TEMPALGOLABEL}
  \end{figure}
}

\usepackage{listings}
\usepackage{tabularx}

\usepackage{pifont}

\begin{document}
\title{A Reasoning Engine for the Gamification of Loop-Invariant Discovery}

\author{Andrew Walter \and Seth Cooper \and Panagiotis Manolios} 
\institute{Northeastern University
  \email{\{walter.a, se.cooper, p.manolios\}@northeastern.edu}}

\authorrunning{Walter, Cooper, Manolios}

\maketitle

\begin{abstract}
  We describe the design and implementation of a reasoning engine that
  facilitates the gamification of loop-invariant discovery. Our
  reasoning engine enables students, computational agents and regular
  software engineers with no formal methods expertise to
  collaboratively prove interesting theorems about simple programs
  using browser-based, online games. Within an hour, players are able
  to specify and verify properties of programs that are beyond the
  capabilities of fully-automated tools. The hour limit includes the
  time for setting up the system, completing a short tutorial
  explaining game play and reasoning about simple imperative
  programs. Players are never required to understand formal proofs;
  they only provide insights by proposing invariants.  The reasoning
  engine is responsible for managing and evaluating the proposed
  invariants, as well as generating actionable feedback.
\end{abstract}

\lstnewenvironment{smalllstlisting}
{\small}
{\normalsize}

\newcommand{\CodeGame}{IDG}
\newcommand{\TraceGame}{\mbox{IDG-T}}
\newcommand{\ISigil}{$[\![$\code{I}$]\!]$}
\newcommand{\GSigil}{$[\![$\code{G}$]\!]$}
\newcommand{\TutorialProgram}{\namef{multiply}}
\newcommand{\ExampleProgram}{\namef{binary-product}}

\newcommand{\InvT}{\code{type-tautology}}
\newcommand{\InvD}{\code{displaced}}
\newcommand{\InvDP}{\code{displaced-pot}}
\newcommand{\InvP}{\code{potential}}
\newcommand{\InvS}{\code{s-inv}}
\newcommand{\InvSN}{\code{s-non-inv}}
\newcommand{\InvI}{\code{inductive}}
\newcommand{\InvN}{\code{non-inv}}

\newcommand{\ProposeProperty}{\namef{Propose\-Property}}
\newcommand{\ProposeLoopInv}{\namef{Propose\-LoopInv}}
\newcommand{\GenCheckVCs}{\namef{GenChkVCs}}
\newcommand{\StmtsBetweenStar}{\namef{StBt*}}
\newcommand{\PrecedingSummaryLocation}{\namef{PreSumLoc}}

\newcommand{\rssa}{RSSA}

\section{Introduction}
\label{sec:intro}
We introduce a reasoning engine that is the key enabling technology of
\CodeGame, the Invariant Discovery Game~\cite{idg}.  The game enables
students and programmers without formal methods expertise to prove
interesting statements about programs. Our reasoning engine does this
by taking possibly incomplete or even incorrect insights about a
program provided by students, programmers or other computational agents,
combining such input with previous inputs, and giving users actionable
feedback highlighting missing insights required to prove correctness.

\begin{figure}
  \centering
  \includegraphics[width=\columnwidth,trim=0in 0in 0in 0in]{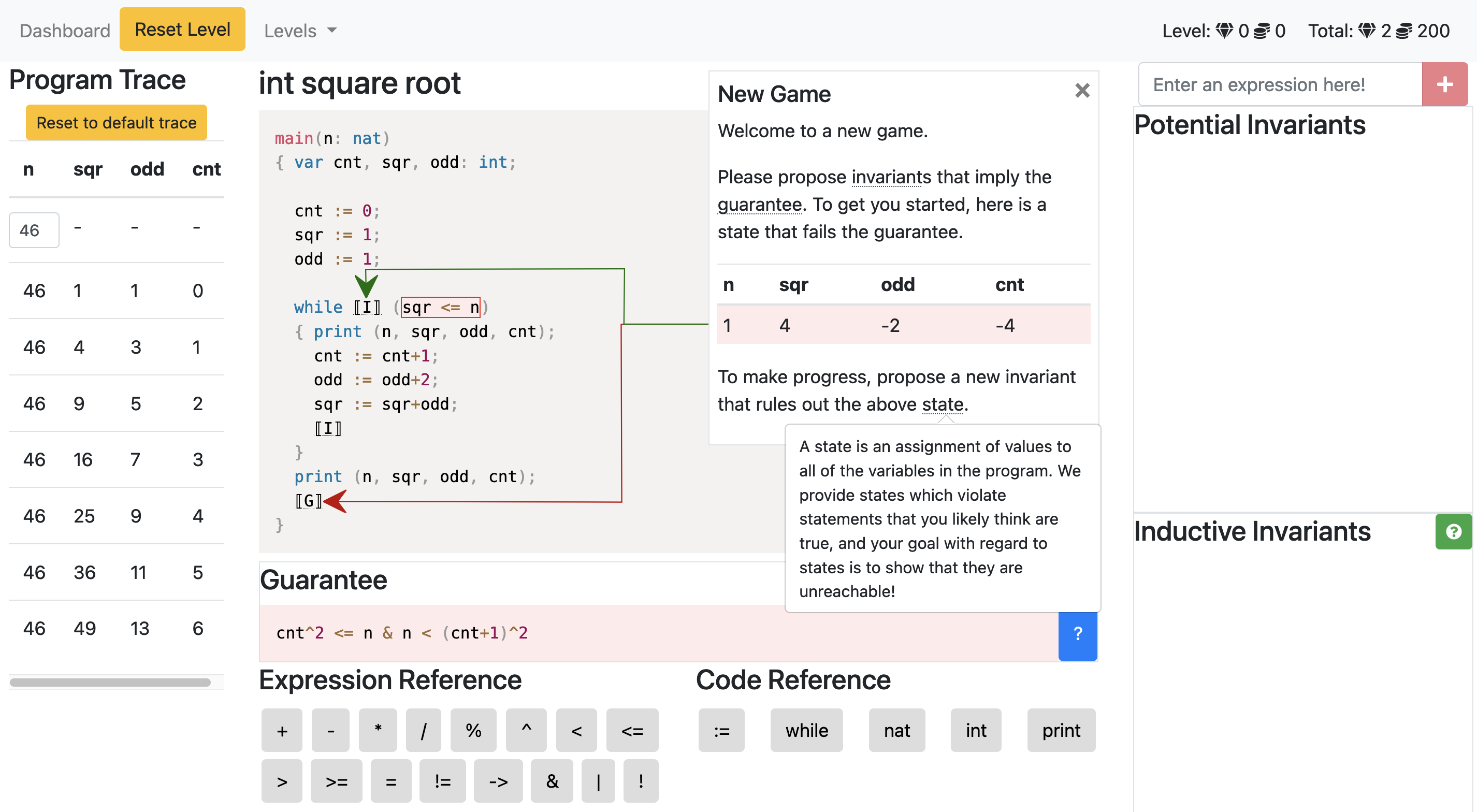}
  \caption{The initial state of the game.}
  \label{fig:int-sqrt-expr1}
\end{figure}

An example of \CodeGame\ is shown in
\figurename~\ref{fig:int-sqrt-expr1}.  To reach this level of the
game, players have to complete a tutorial that explains the interface.
The game can be played at \url{http://invgame.atwalter.com} and the
reader is encouraged to play along.  The program under consideration
is shown in the middle of the figure. The program's guarantee,
\verb!cnt^2 <= n & n < (cnt+1)^2!, is shown near the bottom.  The goal
of the player is to propose enough invariants to enable the reasoning
engine to prove the guarantee.  To help the player, the game provides
feedback that includes a program state which satisfies the proposed
invariants, fails the loop condition and falsifies the guarantee. Such
a state should be unreachable and to make progress, the player is
asked to propose new invariants that rule out this state.

As shown in the figure, players can hover over technical terms such as
\emph{state}, resulting in a pop-up window explaining the
term. Players can also generate program traces by providing inputs; in
the figure a trace for \code{n = 46} is shown.  This trace suggests
that the variables \code{odd} and \code{cnt} can never be negative,
leading to the player proposing \code{odd >= 1} and \code{cnt >= 0} in
the expression box. The player also realizes that \code{odd} must be
odd and proposes \verb!odd % 2 = 1!, after consulting the expression
reference located at the bottom of the screen.  As the player types a
proposed invariant, the game checks that it is satisfied by the
current trace; rows that satisfy the invariant are colored green and
rows that do not satisfy the invariant are colored red (not
shown). Only when all rows are green is the player allowed to propose
the invariant by clicking on the red \code{+} button.

\begin{figure}[t]
  \centering
  \includegraphics[width=\columnwidth]{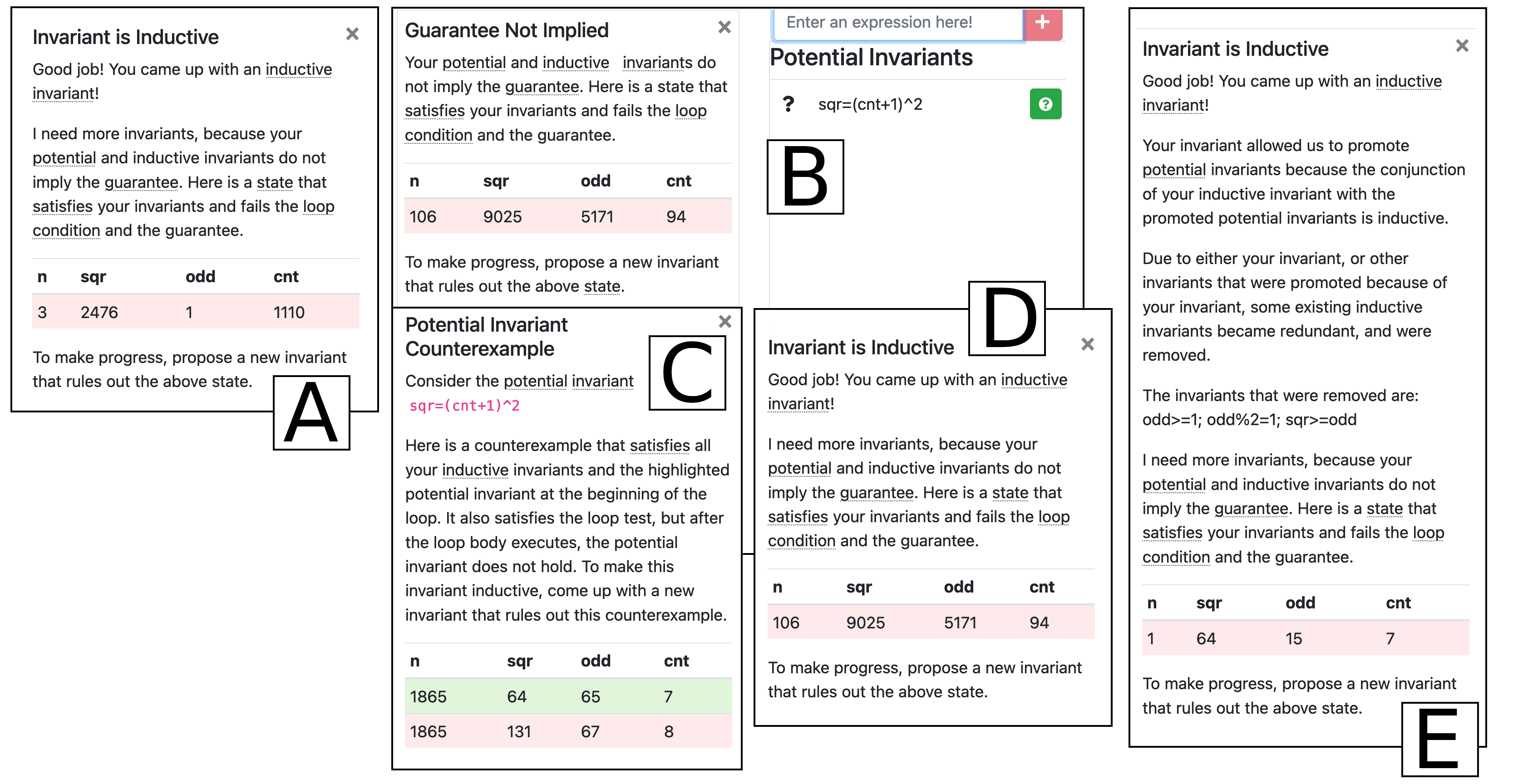}
  \caption{Feedback generated in response to proposed invariants.}
  \label{fig:int-sqrt-expr1collage}
\end{figure}

\CodeGame\ adds the three expressions to the list of inductive
invariants and responds with the feedback shown in
\figurename~\ref{fig:int-sqrt-expr1collage}a, which includes a new
state that the player should rule out by suggesting more invariants.
At this point, the player generates a trace for \code{n = 3}, to
compare an actual trace with the state shown by the game. After trying
a few more traces (consider the trace in
\figurename~\ref{fig:int-sqrt-expr1}), the player recognizes that
\code{sqr} is always the \code{cnt+1}$^{th}$ perfect square, and
enters the expression \verb!sqr = (cnt+1)^2!.  \CodeGame\ then adds
this expression to the list of potential invariants and responds with
the feedback shown in \figurename~\ref{fig:int-sqrt-expr1collage}b. To
explore why this is a potential invariant and not an inductive
invariant, the player clicks on the \code{?} icon next to the
potential invariant resulting in the feedback shown in
\figurename~\ref{fig:int-sqrt-expr1collage}c.  Note that the first
state satisfies the known inductive and potential invariants, but
after executing the body of the loop once, we wind up with the second
state, which does not satisfy the potential invariant.  The problem is
that the player has not (yet) suitably constrained \code{odd}, which
can never be greater than \code{sqr}, so the player proposes
\code{sqr>=odd}, which \CodeGame\ proves to be inductive.
\CodeGame\ responds with the feedback shown in
\figurename~\ref{fig:int-sqrt-expr1collage}d.  After looking at some
traces, the player has the insight \code{odd=cnt*2+1}.
\CodeGame\ confirms that this is an inductive invariant and responds
with the feedback shown in
\figurename~\ref{fig:int-sqrt-expr1collage}e.  Notice that the
potential invariant was promoted to an inductive invariant.  The
promotion highlights one of the ways in which the reasoning engine
manages and uses incomplete information.  In addition, the reasoning
engine also determined that three of the inductive invariants are now
redundant and, in order to limit the cognitive load on users, these
redundant invariants were removed.  The reader is encouraged to finish
the game, which only requires one more invariant.

Notice that players are only responsible for proposing invariants, not
for any kind of formal reasoning.  \emph{Collaborative play} is also
possible by combining proposed invariants from multiple players, some
of whom are humans and some of whom are computational agents. In
addition, the game has been used at Northeastern University to teach
undergraduate students about loop invariants for imperative programs.

An evaluation of \CodeGame\ performed using players from Amazon's
Mechanical Turk showed that \CodeGame\ players were able to
identify invariants that imply correctness for programs beyond the
reach of fully automated systems and that \CodeGame\ was more
effective at eliciting such invariants than previous games~\cite{idg}.

The major contribution of this paper is the design of a reasoning
engine that allows students and programmers without formal methods
knowledge to collaboratively prove statements about programs which
were not provable by best-in-class automated tools~\cite{invgame},
with minimal training, in under one hour.  We describe the core
algorithms for taking incomplete proposed invariants from multiple
sources and combining them to verify programs and to produce concrete
and actionable feedback enabling users to propose further invariants,
without having to understand formal proofs.  Feedback is designed to
leverage the users' existing expertise in programming, \eg, by showing
program traces, exhibiting program states that need to be ruled out by
new invariants and by highlighting relevant portions of the program
code.

\section{Related Work}
There has been significant work in the area of program verification
tools. We give a short overview of several classes of program
verification tools and discuss their relevance to our work. We also
discuss previous work in the area of gamification applied to program
verification.

General-purpose interactive theorem provers are one class of such
tools. Examples include ACL2, ACL2s, Agda, Coq, HOL, HOL-Light,
Isabelle and PVS~\cite{acl2,acl2s,acl2s11,coq,hol,hol-light,isabelle,pvs}.  In
the hands of experts, many amazing theorems have been proved about
complex systems using interactive theorem
provers~\cite{flyspeck,autogodel,fourcolor,acl2commercial,centauracl2}.
These tools can be thought of as proof checkers because users are
responsible for sketching proofs and providing enough guidance so
that the tools accept their proofs.  Since program verification is an
undecidable problem, to be fully general, current tools have to be
interactive and one of the main goals in the area is to provide useful
mechanisms that increase automation. For example in ACL2, there are
dozens of ways of using and combining libraries, theorems, external
tools, specialized reasoning engines, decision procedures,
user-configured proof-search methods, etc. to program the theorem
prover so that it can effectively reason about problems in a
particular domain.  While significant progress has been made, to
effectively use such tools requires significant training over the
course of several months and requires understanding the underlying
language, logic, proof theory and features of the interactive theorem
prover.

Fully automated theorem provers, such as Alt-Ergo, CVC, Inez and
Z3~\cite{alt-ergo,cvc4,inez,z3} are able to automatically prove or
disprove conjectures, but only for limited fragments of logic.
Program verification tools like Spec\#, VCC, and
ESC/Java~\cite{specsharp, vcc, escjava} allow programmers to annotate
their programs with properties that an automated tool then attempts to
prove hold. These tools have numerous options
and significant learning curves.

Outside of the traditional sphere of formal methods research, research
into the gamification of program verification and theorem proving is
an emerging area of research.  The DARPA Crowd Sourced Formal
Verification (CSFV) program, launched in the early 2010s, produced
games which target different kinds of formal verification
problems~\cite{dean2015lessons}. Two of the games developed for the
CSFV program are of particular interest: Xylem and Binary Fission.

In 2018, Bounov \etal\ developed a loop invariant discovery game in
which, unlike many of the CSFV games, users are exposed to
mathematical symbols and operators directly~\cite{invgame}. This game,
called InvGame, is similar to many of the CSFV games in that it does
not display the code being reasoned about. Bounov \etal\ found that,
by crowdsourcing the game on Amazon Mechanical Turk, players
collectively were able to find the loop invariants needed to prove 10
of 14 benchmarks which leading automated program verification tools
could not prove. Bounov \etal\ found that the decision not to abstract
away the underlying math enabled players to use their existing
mathematical expertise while playing the game. On the other hand,
while the choice to not display program code reduced the amount of
information players had to work with, this resulted in a game that
presented a significantly lower cognitive burden for players. Bounov
\etal\ theorize that this made the game easier for non-experts to
play.

We developed the loop invariant discovery game \CodeGame, discussed in
the introduction, which shows code and does not abstract away the
underlying mathematics of the loop invariants, thereby leveraging the
existing programming expertise of players. The evaluation of
\CodeGame\ shows that it is the most effective invariant discovery
game~\cite{idg}.  Quiring and Manolios have recently developed GACAL,
a tool that can play the loop-invariant discovery game, outperforming
most human players and providing an example of a computational agent
that can augment human players~\cite{quiring2020gacal}.

\section{System Architecture}
\label{sec:architecture}

\begin{figure}
  \centering
  \includegraphics[width=7cm]{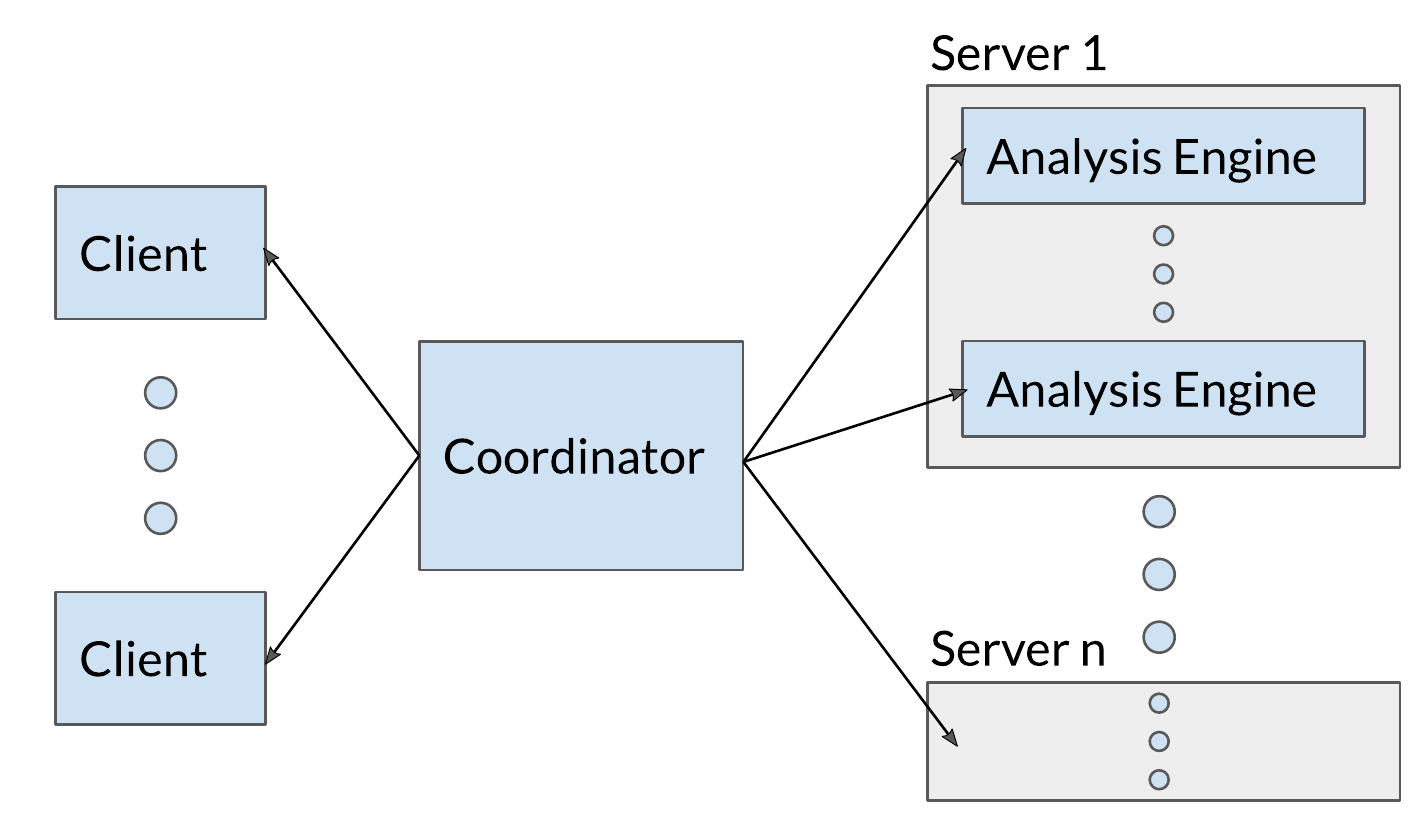}
  \caption{System architecture for \CodeGame}
  \label{fig:architecture}
\end{figure}

The architecture for \CodeGame, shown in
\figurename~\ref{fig:architecture}, consists of clients, a
coordinator, and Invariant Analysis Engine (IAE) instances.  Multiple
clients and IAE instances are allowed, each of which may be on a
separate computer. Players run the client, which handles the UI (User
Interface) and game state, in their browser. When a player suggests an
expression within the client, the client queries the coordinator,
which selects an idle IAE instance and asks it to analyze the
expression. When the IAE instance completes its analysis, the
coordinator forwards the results to the client, which updates the UI
and game state appropriately.

The client is responsible for storing the user's game state, as well
as handling the game UI. For each program the user has attempted, the
client persists a record of the expressions the user proposed, even
those which are not shown in the UI. This allows the user to continue
working on programs even after leaving the game. The client also
computes user scores based on the expressions they submit.

The client performs local checks to prevent the user from submitting
obviously useless expressions to the IAE. This includes ensuring that
expressions are syntactically valid, different from previously
submitted expressions and not falsified by the current trace.  These
checks are relatively simple and implementing them in the client, as
opposed to having to engage the IAE, results in instantaneous user
feedback.

The coordinator is responsible for choosing an IAE instance to handle a
client query, as well as storing data logged by clients and
maintaining a cache of the programs that users submit.
The coordinator is intended to be low-overhead, with as much of the
heavy computation limited to the client and the IAE instance. However,
should multiple coordinators be necessary to handle the volume of
queries being made by clients, the design of the coordinator does not
preclude this.

The IAE, which is the focus of this paper, is described in 
Section~\ref{sec:iae}.

\section{Invariant Analysis Engine}
\label{sec:iae}

The playthrough of \CodeGame\ described in the introduction
highlighted three different broad types of actions that the Invariant
Analysis Engine (IAE) must perform: (1) trace generation, (2)
expression characterization, handling and promotion and (3)
counterexample and feedback generation.

The above actions are performed on arbitrary programs written in the
Simple Imperative Programming language (SIP), whose abstract syntax is
shown in Table~\ref{tbl:syntax}.  SIP expressions consist of Boolean
and arithmetic expressions over unbounded integers, natural numbers,
and rational numbers, as well as function calls.  Executable
statements include variable declarations, assignments, compositions,
conditionals, while loops and print statements.  SIP also includes
assumption, assertion and \code{cassign} assignments:
\code{cassign([$\mathit{vars}$], $\phi$)} nondeterministically assigns
values to the variables in $\mathit{vars}$ that satisfy the expression
$\phi$.  Function definitions optionally include \code{pre} and
\code{post} statements, corresponding to pre- and post-conditions.
The semantics of SIP programs are now standard.

\begin{table}
  \caption{Abstract Syntax of SIP}
  \label{tbl:syntax}
  \ \ $\langle r \rangle$ indicates that $r$ is optional and
  $r\ldots$ indicates 0 or more $r$'s.\\
  \begin{tabularx}{\textwidth}{ l @{\hspace{1\tabcolsep}} c @{\hspace{1\tabcolsep}} X}
    $i$ & & identifier \\
    $t$ & & type specifier (Boolean, Natural, Integer, Rational) \\
    $e$ & & Boolean or numeric expression \\
    $s$ & ::= & \lstinline!var\ $i$ : $t$ | i:= $e$ | \ $s$; $s$ |
    if($e$)\{$s$\}else\{$s$\} | while$\langle [e] \rangle$($e$)\{$s$\}
    | print($e\ldots$)|!
    \newline
    \lstinline!assume($e$) | assert($e$) | claim($e$) | cassign([$i\ldots$],$e$)! \\
    $f$ & ::= & \lstinline!fn $\ i$ ($i$: $t \ldots$):$t\ $ \{ $\langle$pre($e$); $\rangle\:\langle$ post($e$); $\rangle\:s$ \}!\\
    $p$ & ::= & \lstinline!$f \ldots$!
  \end{tabularx}
\end{table}

Our goal is to enable human and computational agents to
collaboratively reason about SIP programs, but due to space
limitations, in this paper we make the simplifying assumptions that
SIP programs are well-typed and consist of a single function with a
single while loop. All of the levels in \CodeGame\ satisfy these
restrictions.

We defined the syntax and semantics of SIP programs using the ACL2s
theorem prover~\cite{acl2sweb}.  We implemented a trace generation
capability that allows us to generate a trace, given a SIP program as
input.  Finally, we implemented a procedure that given such a SIP
program, generates verification conditions for the loop invariant and
the post condition (referred to as the guarantee in \CodeGame).
  
\subsection{Expression Characterizations}
\label{sec:exprchar}

An \CodeGame\ player's goal is to submit expressions that progressively
strengthen a loop invariant until it is strong enough to prove the
property corresponding to the guarantee.  To support collaborative and
incremental reasoning, expressions are submitted incrementally,
potentially from multiple sources, as the game is played.
The IAE 
determines which proposed expressions provide useful information about
the program under consideration, and which do not, by characterizing
proposed expressions, as described below.

Let $e$ be the expression under consideration.  The characterization
of $e$ depends on $T$, the type information for the program under
consideration. $T$ depends only on the program and is not affected by
game play. The characterization of $e$ also depends on the history of
expression characterizations that occurred during the game play before
the proposal of $e$.  Part of this history is recorded in $I$, the set
of currently known inductive loop invariants, \ie, $I$ is a set of
previously proposed expressions and $\bigwedge I$ is inductive.  $I$
depends on game play; initially $I$ is empty, but as agents play the
game, they propose expressions that may be added to $I$.  We also have
$P$, the of currently known \emph{potential invariants}: expressions
with the potential to be characterized as inductive invariants after
more game play. How this happens will be described shortly.  Given
this setup, we characterize $e$ as follows.

\newcounter{expchar}
\stepcounter{expchar}
\noindent 
(\arabic{expchar}) \InvT: $e$ is a tautology, assuming the program's
types, \ie, $T \Rightarrow e$. For example $y-x \leq y$ is a type
tautology if $x$ is a natural number and $y$ is an integer.  In the
sequel, we assume $T$ implicitly as a hypothesis as appropriate.

\stepcounter{expchar}
\noindent 
(\arabic{expchar}) 
\InvN: $e$ is not an invariant and the IAE has a counterexample $c$,
an assignment to the input variables that satisfies $T$ such that
running the program with $c$ results in a state at the location of the
loop invariant that falsifies $e$. This state may occur when the loop
is first reached, or after some number of loop iterations.

\stepcounter{expchar}
\noindent 
(\arabic{expchar}) \InvI: $e$ is inductive, \ie, for all legal inputs,
$e$ holds when execution of the program first reaches the loop and for
all states in which $I$, $e$ and the loop test hold, then after
executing the body of the loop, $e$ holds. Notice the dependence on
$I$, the set of currently known inductive loop invariants.
Note that
all \InvT\ expressions are also \InvI\ expressions, so the
characterizations are not disjoint.

\stepcounter{expchar}
\noindent 
(\arabic{expchar}) \InvP: $e$ is a potential invariant if it is not
\InvN\ and it is not \InvI. For example, if $e$ is an invariant, but
is not inductive given $I$, then $e$ will be characterized as being
potential. Notice the dependence on $I$. If $I$ is expanded in future
game play, it is possible that an element $p \in P$ becomes inductive,
with respect to the expanded $I$, in which case we say that $p$ can be
\emph{promoted}.

If $e$ is inductive, we can add it to $I$ and if it is potential, we
can add it to $P$, but we have to be careful because any of these
moves can have consequences, which lead to two more characterizations.

\stepcounter{expchar}
\noindent 
(\arabic{expchar}) \InvD: $e$ is a displaced invariant if it is
implied by the existing set of invariants ($I \Rightarrow e$).
Displaced invariants are redundant, as they provide no new
information beyond what is already known, so they can be dismissed.

\stepcounter{expchar}
\noindent 
(\arabic{expchar}) \InvDP: $e$ is a displaced potential invariant if
it is equivalent to some subset of potential invariants, assuming that
the set of invariants hold. That is, $\langle \exists S \subseteq P ::
I \Rightarrow (\bigwedge S = e) \rangle$.  Displaced potential
invariants provide no new information and also do not limit future
promotions, so they can be dismissed.  Not limiting future promotions
is the reason why the definition of a displaced potential invariant is
stricter than the definition of a displaced invariant.  Due to space
limitations, we ignore this characterization in the sequel.

We note that the characterization problem is undecidable, as we have
to check the validity of formulas that include addition and
multiplication, hence, our final characterization.

\stepcounter{expchar}
\noindent 
(\arabic{expchar}) \Unknown: When none of the above cases hold,
we say that $e$ is \Unknown.

\subsection{Theorem Prover}

The IAE depends on a theorem prover that provides the functionality we
describe in this section.  We use the ACL2s theorem prover, but any
appropriate theorem prover can be used.  However, we note that the
\texttt{cgen} counterexample generation~\cite{cgen} and
\texttt{defdata} data definition~\cite{defdata} features of ACL2s are
key technologies in our implementation of the IAE.

The logic of the theorem prover must be expressive enough to support
the Verification Conditions (VCs) generated by the IAE, which include
implicitly universally quantified formulas over the SIP-supported
types (Booleans, natural numbers, integer and rationals) and operators
(arithmetic and Boolean).  The characterizations from the previous
section are examples of the kinds of queries the IAE generates.

Given a query, the theorem prover either returns the constant \Proved,
if it can prove that the formula holds, or it returns counterexamples
disproving the query, or \Unknown\ is returned. Due to undecidability,
we cannot rule out the last case.

The IAE interacts with the underlying theorem prover through a single
procedure, \GenCheckVCs. This procedure generates VCs for the given
SIP program, using the verification condition generation capability
mentioned above, and submits the VCs to the underlying theorem prover.
In more detail, \GenCheckVCs\ takes as input a SIP program, $p$, and a
sequence of SIP statements, $t$, over the variables in $p$, with
exactly one \code{assert} statement that must be the last element of
$t$.  The procedure generates verification conditions that check
whether the assertion in $t$ holds, under the assumption that the
variables appearing in $t$ have the types specified in $p$.  Finally,
we implemented a procedure that given such a SIP program, generates
verification conditions for the loop invariant and the post condition
(referred to as the guarantee in \CodeGame).

If an algorithm listed here does not explicitly handle the \Unknown\
case when proving a VC, assume that the IAE will notify the client
that it couldn't generate any useful feedback for the user.

\subsection{Core IAE Algorithms}

We now present the core IAE algorithms.
We use the notation \textbf{alias} \varf{foo} \assign \varf{bar.baz}
inside of algorithms to mean that \varf{foo} is just another name for
\varf{bar.baz}, \ie, assigning a value to \varf{foo} will also
modify the value of \varf{bar.baz}, and vice versa.

The data structures used to store expressions, expression states, and
programs are defined in \figurename~\ref{lst:typedefs}.  For
presentation purposes, only a small subset of the data structures and
fields in the IAE are shown, \eg, the IAE includes data structures and
fields to support the recording of all queries and characterization
results, which are useful for evaluating gameplay mechanics. In
addition, we have simplified what actually happens, \eg, in this paper
we only consider programs with a single while loop, which
significantly simplifies matters. In the actual implementation, we
have a summary-based approach that supports reasoning about arbitrary
SIP programs.

\begin{myalgorithmic}{Type definitions for IAE}{lst:typedefs}
  \Structure{InvariantState}
    \field{IInvs}{inductive loop invariants}
    \field{PInvs}{potential invariants}
  \EndStructure
    
  \Structure{Program}
    \field{Test}{The test for the loop}
    \field{Body}{The body of the loop}
  \EndStructure
\end{myalgorithmic}

Commonly occurring arguments to the procedures defined in this section
include: (1) \varf{p}, the type-checked SIP program under
consideration, (2) \varf{s}, a structure encapsulating the current
state of gameplay as described above and (3) \varf{e}, the expression
under consideration, which is a Boolean expression over program
variables.  If a set of expressions is used in a Boolean context, then
it is is implicitly conjoined.

Table~\ref{tab:functionlisting} contains information about the
procedures that are used in the description of the IAE algorithms
below and is useful as a reference for understanding the algorithms in
this paper.

\begin{table*}[tb]
  \centering
  \caption{IAE Procedure Listing}
  \label{tab:functionlisting}
  \begin{tabular}{|p{2.8cm}|p{9cm}|}
    \hline
    \GenCheckVCs\newline(\varf{p}, \varf{g}) & Generate VCs that the SIP
    statement \varf{g} gives rise to, with type information from program \varf{p}.
    Returns \Proved\ if VCs are proved, a
    counterexample if one was found and \Unknown\ otherwise.\\
    \hline
    \ProposeLoopInv\newline(\varf{p}, \varf{s}, \varf{e}) &
    Characterizes \varf{e} and updates any other expression
    characterizations affected. Returns (\varf{kind}, \varf{res},
    \varf{s}) where kind is the expression characterization for
    \varf{e}, \varf{res} is information that may be used for feedback,
    and \varf{s} is the updated game state.\\ 
    \hline
    \namef{HasCtrxs}(\varf{c}) &  Returns \True\ if \varf{c}
    corresponds to a theorem prover query that found counterexamples
    and \False\ otherwise.\\
    \hline
    \namef{GenTrace}(\varf{p}, \varf{a}) & Generates a trace
    of $p$ using variable assignments from 
    \varf{a}.\\ 
    \hline
    \namef{CheckUptoLoop}\newline(\varf{p}, \varf{e}) &
     Generates subprogram of $p$ up to, but not including the
     loop, ending with a  statement asserting $e$.\\
    \hline
    \namef{CheckLoop}\newline(\varf{p}, \varf{e}) &
    Generates subprogram of $p$ up to 
    the loop, which is modified to allow execution to
    nondeterministically terminate early, and followed by
    a statement asserting $e$.\\
    \hline
    \namef{Promote}(\varf{p}, \varf{s}) & Promotes as many
    potential invariants as possible to inductive invariants, for
    program $p$ and state $s$.\\
    \hline
    \namef{RemDisplaced}(\varf{s}) & Removes displaced
    inductive invariants from  \varf{s}. \\ 
    \hline
  \end{tabular}
\end{table*}

The procedure for evaluating proposed loop invariants is shown in
\figurename~\ref{lst:proposeloopinv}.  The procedure takes as input
(1) $p$, the program under consideration, (2) $s$, the structure
described above, which includes the set of inductive and potential
invariants and (3) $e$, the proposed invariant. It returns a tuple
(\varf{kind}, \varf{res}), where \varf{kind} is the characterization
of $e$ and \varf{res} includes extra information when the
characterization is \InvN: this extra information includes a trace
that can be used to show the player why $e$ is not an invariant.  The
procedure also updates $s$; this happens in the call to procedure
\namef{Promote} and only when the characterization of $e$ is \InvI\ or
\InvP.

\begin{myalgorithmic}{ProposeLoopInv}{lst:proposeloopinv}
  \Procedure{ProposeLoopInv}{\varf{p}, \varf{s}, \varf{e}}
  \LeftComment{Returns a tuple (\varf{kind}, \varf{res})}
  \Alias{I}{s.IInvs}
  \Alias{P}{s.PInvs}
  \State \varf{tautology} \assign
     \GenCheckVCs(\varf{p}, \code{assert(\varf{e});})
  \If{$\varf{tautology} = \code{proved}$}
    \State \Return (\InvT, \code{nil})
  \EndIf
  \State \varf{displaced} \assign
     \GenCheckVCs(\varf{p}, \code{assume(\varf{I}); assert(\varf{e});})
  \If{$\varf{displaced} = \code{proved}$}
    \State \Return (\InvD, \code{nil})
  \EndIf
  \State \varf{eChk} \assign
         \GenCheckVCs(\varf{p}, \namef{CheckUptoLoop}(\varf{p}, \varf{e}))
  \If{$\namef{HasCtrxs}(\varf{eChk})$}
    \State \varf{trace} \assign \namef{GenTrace}(\varf{p}, \varf{eChk})
    \State \Return (\InvN, (\varf{eChk}, \varf{trace}))
  \EndIf
  \If{$\varf{eChk} = \Unknown$}
    \State \Return (\Unknown, \code{nil})
  \EndIf
  \State \varf{eChk} \assign
         \GenCheckVCs(\varf{p}, \namef{CheckLoop}(\varf{p}, \varf{e}))
  \If{$\namef{HasCtrxs}(\varf{eChk})$}
    \State \varf{trace} \assign \namef{GenTrace}(\varf{p}, \varf{eChk})
    \State \Return (\InvN, (\varf{eChk}, \varf{trace}))
  \EndIf
  \State $P$ \assign $P \cup \{e\}$  
  \State \namef{Promote}(\varf{p}, \varf{s})
  \If{$e \in I$}
    \State \Return (\InvI, \nil)
  \Else \State \Return (\InvP, \nil)
  \EndIf
  \EndProcedure
\end{myalgorithmic}

\namef{ProposeLoopInv} starts by checking whether $e$ is a \InvT, by
querying the theorem prover. The call to \GenCheckVCs\ will generate a
logical formula stating that $e$ follows from the just the type
information in program $p$. Notice that in this call to \GenCheckVCs,
we have satisfied the preconditions for the procedure as the second
argument is a sequence of SIP statements over the variables in $p$,
with exactly one \code{assert} statement, which is the last statement
in the sequence. We will leave such checks to the reader in the
sequel. The next check is whether $e$ is displaced, which corresponds
to checking if $e$ holds, assuming $I$, the set of invariants and the
type information in program $p$. Next, we check if $e$ holds the first
time program execution reaches the loop, using procedure
\namef{CheckUptoLoop}, which takes as input a program, $p$, and an
expression, $e$, and generates a sequence of SIP statements similar to
$p$. The sequence includes all statements in $p$ up to, but not
including the loop. Variable declarations are also not
included. Lastly, we add a final statement asserting $e$.  If we find
counterexamples, then clearly $e$ is not a loop invariant. If the
theorem prover returns \Unknown, then we return.  Otherwise, we know
that $e$ holds when program execution reaches the loop for the first
time. Notice that it is still possible for $e$ to not be an invariant,
and the next check is our final attempt to check this. We check if $e$
is a \InvN, using the procedure \namef{CheckLoop}, which takes as
input a program, $p$, and an expression, $e$, and generates a sequence
of SIP statements similar to $p$. The sequence includes all statements
in $p$ up to the loop, but not including variable declarations. In
addition, the loop is modified to allow execution to
nondeterministically terminate early. Lastly, we add a final
statement asserting $e$.  ACL2s allows us to express such formulas,
using recursion, but if the underlying theorem prover is not
expressive enough, it would be fine to either skip this test or to
traverse the loop no more than some constant number of times.
If no counterexamples are found, then we will treat $e$ as a 
potential invariant, by adding it to $P$ and the \namef{Promote}
procedure will be called to check if $e$ or any other element of $P$
can be promoted. We note that it is possible that $I \wedge e$ is not
inductive, but for $e$ and other expressions in $P$ to be promoted
anyway.

The procedure for promotion is shown in \figurename~\ref{lst:promote}.
The procedure takes as input $p$, the program under consideration and
$s$, the structure described above, which includes the set of
inductive and potential invariants.  It modifies $s$ by updating the
potential and inductive invariants in $s$.

\begin{myalgorithmic}{Promote}{lst:promote}
  \Procedure{Promote}{\varf{p}, \varf{s}}
  \LeftComment{Promotes PInvs and updates \varf{s}}
  \Alias{P}{s.PInvs}
  \Alias{I}{s.SInvs}
  \State \varf{lt} \assign \varf{p}.\varf{test}
  \State \varf{lb} \assign \varf{p}.\varf{body}
  \State \varf{X} \assign \varf{P}
  \Repeat
    \State \varf{Done} \assign \True
    \For{$\varf{x} \in \varf{X}$}
      \State \varf{g} \assign \code{assume($\varf{I}
        \wedge \varf{X} \wedge \varf{lt}$);
        $\varf{lb}$; assert($\varf{x}$);}
      \If{\GenCheckVCs(\varf{p}, \varf{g}) $\neq \Proved$}
        \State \varf{X} \assign $\varf{X} \setminus \{\varf{x}\}$
        \State \varf{Done} \assign \False
      \EndIf
    \EndFor
  \Until{$\varf{Done}$}
  \State $\varf{I} \assign \varf{I} \cup \varf{X}$
  \State $\namef{RemDisplaced}(\varf{s})$
  \State $\varf{P} \assign \varf{P} \setminus \varf{X}$
  \EndProcedure
\end{myalgorithmic}

\namef{Promote} starts with some simple assignments, including
initializing $X$ to be a copy of $P$, the potential invariants in $s$.
The outer \code{repeat} loop is taken as long as an element is removed
from $X$, which can only occur $|X|$ times.  Each iteration of the
\code{repeat} loop has an inner \code{for} loop that can potentially
remove elements from $X$. The key insight is that an element $x$ of
$X$ is not inductive if it does not hold after the body of the loop,
assuming that $I, X$ and the loop test hold.  This check is repeated
until we reach a fixpoint. That is, when the \code{repeat} loop
finishes, we know that assuming $I, X$ and the loop test all hold,
then after execution of the loop body, $I, X$ hold. That $I$ holds
follows from the assumption that $\mathit{s.PInvs}$ is a set of loop
invariants, so $I$ holds after the loop with the weaker hypothesis
that $I$ holds before the loop.  To see that $X$ holds after the loop,
suppose not, then there is some $x \in X$ which does not hold and it
would have been removed by the \code{for} loop. Notice that $X$ is the
largest subset of $P$ which can be promoted. The final value of $X$ is
determined by a greatest fixpoint computation. Note that if we use a
least fixpoint computation, by adding elements of $P$ to $I$ one at a
time, every element so promoted is really an invariant, but we are not
guaranteed to have promoted all the potential invariants that are
promotable. The procedure $\namef{RemDisplaced}$ maintains the
invariant that $I$ does not include any invariant that is implied by
the rest of the invariants.  When we update $I$ by adding $X$ to it,
we open up the possibility that $I$ contains displaced invariants,
\eg, it is possible that $X$ contains a strong invariant that strictly
implies multiple invariants in $I$. Hence, $\namef{RemDisplaced}$
takes $s$ as input and removes displaced invariants from $I$.
As mentioned previously, IAE also performs displacement for
potential invariants, which is a more complicated process.

\subsection{IAE Generated Feedback}

The IAE provides useful feedback to the user. We have already seen how
this works if the proposed invariant is characterized as a tautology,
a displaced invariant or a non-invariant. If neither of the cases
hold, then the IAE check whether the \CodeGame\ level has been
\textit{solved}: the program's guarantee holds when program execution
reaches the guarantee starting from any state that satisfies all of
the known inductive invariants, the negation of the loop test and
executes all the code following the loop.

If any promotion occurs, then the IAE will generate a query to
check if the level has been solved. If so, the user is congratulated
and is able to move on to another level. If not, any generated
counterexamples are recorded.

If the level has not been solved, then the IAE will generate a query
just like the solved query, except starting states are assumed to
satisfy all of the known inductive \emph{and potential} invariants.
If a counterexample to this query is found, the IAE will return it to
the user and the user will be asked to propose an invariant that rules
out this state. Notice that such a counterexample illustrates that
more invariants are needed, even if we consider the potential
invariants to be inductive. If no counterexample is found to this
query, the IAE will use the counterexamples generated from the solved
query.

The user is also able to request interesting counterexamples,
including counterexamples to the inductivity of potential invariants.

\section{Evaluation}

In order to evaluate the effectiveness of the reasoning engine, we
recruited 300 players from Amazon's Mechanical Turk platform and
randomly assigned them to play \CodeGame\ and \TraceGame, a game based
on InvGame~\cite{invgame}, the state-of-the-art in invariant discovery
games. \TraceGame\ does not show code; it only shows traces. We
implemented \TraceGame, instead of using data from
InvGame~\cite{invgame}, in order to limit confounding variables, such
as differences in player populations, user interface elements,
definitions, metrics and statistical tests.

In a previous paper, we evaluated the gamification aspects of the data
collected, which included the experimental setup, the provenance of
the benchmark problems used, information on the participants, a
comparison of levels solved, cheesability considerations, player skill
data and player feedback~\cite{idg}.  A very brief, incomplete,
summary of the previous evaluation is provided. The games consisted of
twelve levels.  For eleven of these levels, there was at least one
\CodeGame\ player who was able to (individually) solve the level.  For
\TraceGame, only seven of the levels were solved
individually. However, if all the proposed invariants by all the
\TraceGame\ players are collected, then a total of nine levels were
solved. The level that was not solved by \CodeGame\ players was also
not solved by \TraceGame\ players.

For all analyses below, we only consider data for users who completed
at least one level. The first level players were asked to solve was
exactly the same as the level used in the tutorial, hence we used this
criterion to remove players who were gaming the game. For all of the
experiments reported, we identified the metrics and p-values (0.05)
that we consider statistically significant before running the
analyses.

\begin{figure}
  \centering
  \includegraphics[width=0.8\textwidth]{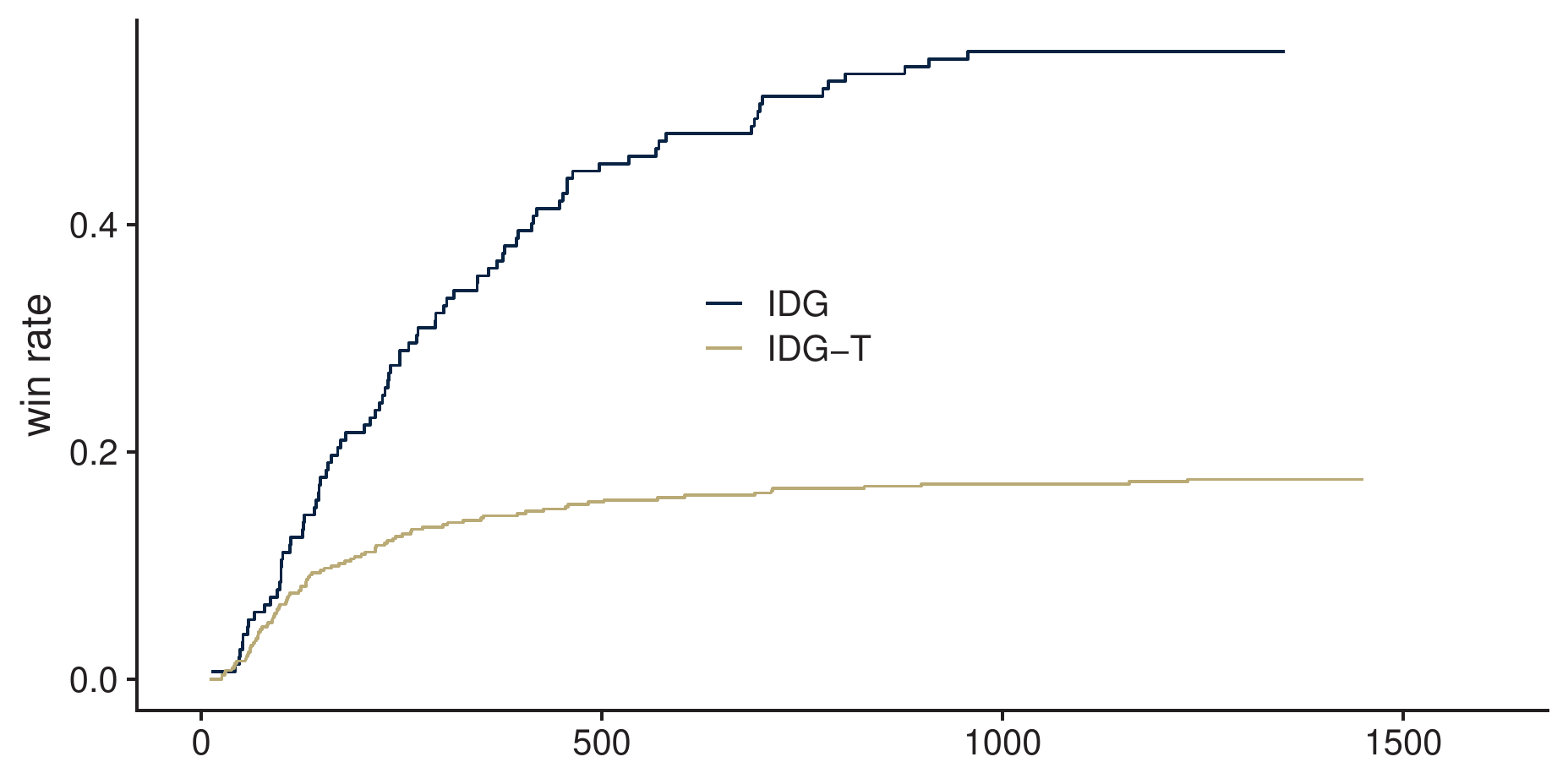}
  \caption{The win rate for
    \CodeGame\ and \TraceGame\ versus active play time, in seconds. }
  \label{fig:win-plot}
\end{figure}

In Figure~\ref{fig:win-plot}, we show the \emph{win rate} of players
for both games.  The $x$-axis corresponds to time in seconds and the
$y$-axis is the win rate of the players.  Each player attempted some
number of levels, typically six; the attempted levels for a game is
the sum of the attempted levels over all the players of the game. The
\emph{win rate} of a game at time $t$ is the ratio of levels solved to
levels attempted, after $t$ seconds of game play. Notice that the
numerator in the ratio depends on $t$, but the denominator does not,
\eg, if at time 500, \CodeGame\ players solved 1,000 levels out of a
total of 2,000 attempted levels, then the win rate, at time 500, is
0.5.  What the graph shows is that at 1,500 seconds, the
\CodeGame\ win rate is 0.553, while the \TraceGame\ win rate is only
0.176. The win rate allows us to estimate the expected number of
levels players of the games will solve within a certain amount of
time. As the figure clearly shows, after any non-trivial amount of
gameplay, \CodeGame\ players solve more levels than
\TraceGame\ players.

Care was taken to deal with outliers.  For example, we noticed that
certain players loaded levels, but did not attempt to solve
them. Therefore, only players who proposed an expression for a level
and had a total active level play time of more than 10 seconds were
considered to have attempted a level. No level was solved in less then
20 seconds. We also noticed that four players spent more than 1,500
seconds on a level, but when we looked at the logs, there were
frequent periods of inactivity, indicating that they were multitasking
or distracted, hence we ignored these players.  The computation of
active level play time is somewhat complex. If a user played the same
level in multiple browser sessions, we include time spent on all plays
and added a 30 second penalty for each session. We ignored periods of
inactivity that lasted at least 5 minutes, but added a 2 minute
penalty for each such period.  This allows us to better approximate
the actual active time spent by players, \eg, there were players with
total, unadjusted play time on a single level of over 4,000 seconds,
almost all of which was idle time.

We use the Mann-Whitney $U$ test to compare the win rates over time
for \CodeGame\ and \TraceGame. The test indicates a statistically
significant difference between the two curves ($p=1.24$e-$30$).  The win
rate was partly motivated by cumulative incidence, from the field of
survival analysis~\cite{survivalanalysis}.

\begin{figure}
  \centering
  \includegraphics[width=0.8\textwidth]{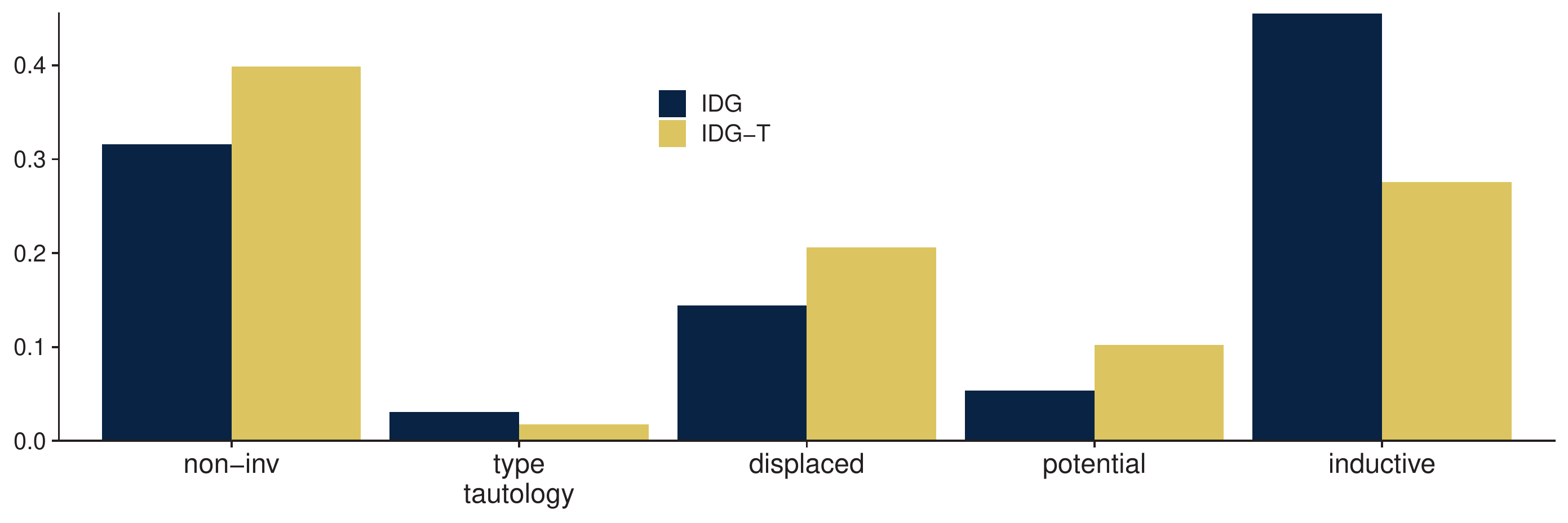}
  \caption{Distribution of the types of submitted expressions, in
    aggregate. Types are ordered from left to right from least useful
    (non-inv) to most useful (inductive).}
  \label{fig:inv_kind_nonstacked}
\end{figure}

\figurename~\ref{fig:inv_kind_nonstacked} sheds light on why
\CodeGame\ players outperformed \TraceGame\ players:
\CodeGame\ players submitted significantly more useful expressions
(inductive and potential invariants) than \TraceGame\ players.  This
was due to the feedback generated by the reasoning engine.
A Pearson's Chi-Squared test with Yates' continuity correction
indicates a statistically significant correlation between game variant
and the distribution of expression types submitted ($p=2.97$e-$24$,
Cramer's $V=0.16$).

\begin{figure}
  \centering
  \includegraphics[width=\textwidth]{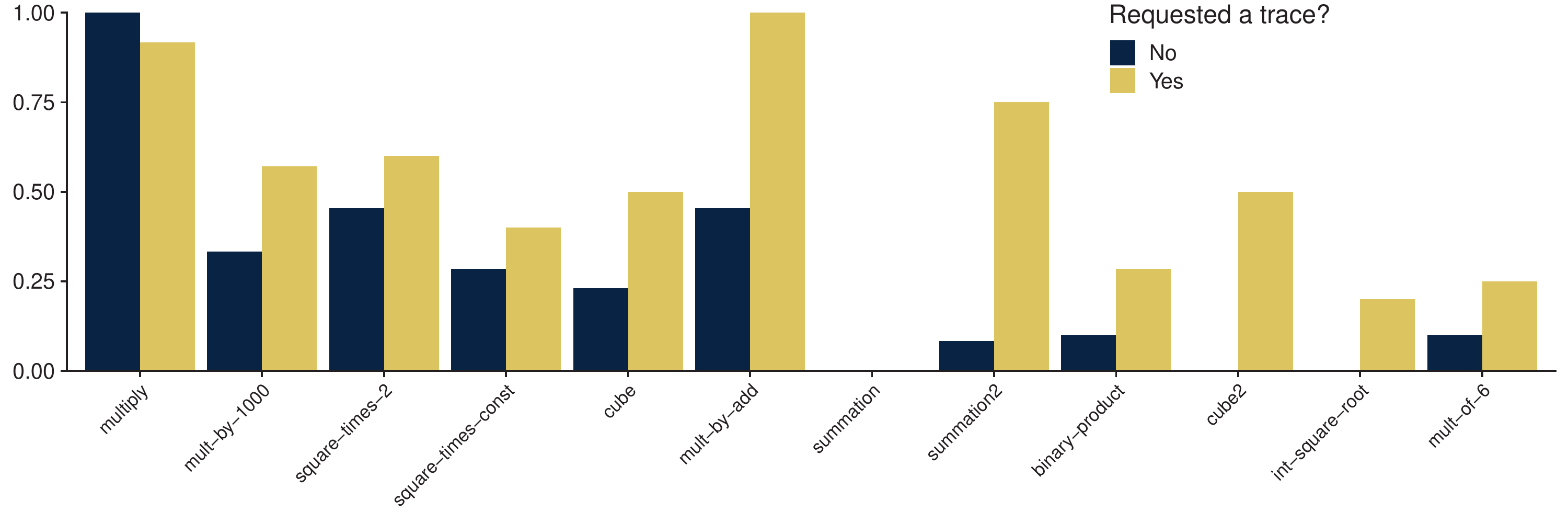}
  \caption{Prove ratio for \CodeGame\ players by level,
    stratified by whether the player requested a trace when
    playing the level. }
  \label{fig:trace_prove_ratio}
\end{figure}

\figurename~\ref{fig:trace_prove_ratio} shows how the \emph{prove
ratio} for \CodeGame\ players, the proportion of players who proved a
level to those who attempted it, differs based on whether players
requested a trace or not. All shown ratios have a non-zero
denominator, \ie, for every attempted level, there was at least one
player who requested a trace and one who did not.  \CodeGame\ provides
the player with a trace when the level is loaded, and may provide
either single states or snippets of traces in its feedback, but the
ability to generate custom traces for given input values seems to be
quite helpful for discovering invariants.  A Pearson's Chi-Squared
test with Yates' continuity correction indicates a statistically
significant correlation between whether a participant proved a level
and whether they requested a trace ($p=0.046$, Cramer's
$V=0.14$). Additionally, this feature was used often, as more than
$29\%$ of all \CodeGame\ level attempts used it.

Our evaluation shows that \CodeGame\ is more effective than
\TraceGame\ at eliciting useful invariants from players. The key
enabling technology for \CodeGame\ is the reasoning engine.

\section{Future Work}
There are many interesting research directions to pursue, some of
which we outline in this section.  One idea is to develop more
educational games that can be used to introduce program verification
to students. We already use \CodeGame\ at Northeastern University to
introduce verification condition generation and reasoning about
imperative programs to undergraduate students.  We would like to
create games that support more complicated programs, using richer
languages and moving towards the analysis of industrially-relevant
programs.  Also, we would like to provide greater support for
crowdsourced program verification, more visualization capabilities and
more customizable games. Finally, we believe that there are many
opportunities to develop computational agents that can augment and
complement human cognitive abilities. The GACAL system is an early
example~\cite{quiring2020gacal}.

\section{Conclusion}
\label{sec:conclusion}

We have described the design and implementation of an interactive
reasoning engine for loop-invariant discovery games that enables
regular programmers without formal methods training to collaboratively
prove program correctness of simple imperative programs.  The games
are Web-based, requiring only a browser to use. After a ten minute
tutorial, players with no formal methods expertise are able to specify
and check loop invariants. Our reasoning engine provides actionable
feedback in the form of program states that players are asked to rule
out. This feedback enables players to effectively use their
programming insights to propose invariants and check invariants.
A detailed evaluation has shown that our game is more effective than
existing games in soliciting useful expressions from players and in
helping players prove program correctness.

\section*{Acknowledgments}
Thanks to Bounov, DeRossi, Menarini, Griswold and Lerner for sharing
code for InvGame; the Mechanical Turk users for playing our games; and
our students.

\bibliographystyle{splncs04}
\bibliography{paper}

\end{document}